%% file: root_verify.tex
\documentclass[letterpaper,10pt,conference]{ieeeconf}

\IEEEoverridecommandlockouts
\usepackage{graphics}
\usepackage{graphicx}
\usepackage{mathptmx}
\usepackage{times}
\usepackage{cite}
\usepackage{amsmath}
\usepackage{amssymb}
\usepackage{booktabs}
\usepackage{url}
\usepackage{xcolor}
\newcommand{\SI}[2]{#1\,#2}
\usepackage[hidelinks]{hyperref}

\title{\LARGE \bf
SmellDiffusion: Diffusion-Based Quadruped Navigation with Olfactory Scene Graphs
}

	\author{Faith Ogunwoye$^{*}$, Iana Zhura$^{*}$, Hajira Amjad, Timofei Kozlov, \\ Didar Seyidov, Dmitrii Plotnikov, Fedor Fedorov, and Dzmitry Tsetserukou%
	\thanks{$^{*}$Equal contribution.}%
	\thanks{All authors are with the Skolkovo Institute of Science and Technology, Moscow, Russia. Email: {\tt\small \{faith.ogunwoye, iana.zhura, hajira.amjad, timofei.kozlov, didar.seyidov, dmitrii.plotnikov, fedor.fedorov, d.tsetserukou\}@skoltech.ru}}%
}

\begin{document}

\maketitle
\AddToHookNext{shipout/foreground}{%
  \put(54,-759){\parbox[b]{252pt}{\normalfont\footnotesize
  Research reported in this publication was financially supported by the RScF grant No. 26-19-00828.}}%
}
\thispagestyle{empty}
\pagestyle{empty}

\input{sections/Abstract}

\section{Introduction}
\input{sections/Introduction}

\section{Related Work}
\input{sections/Related_work}

\section{Method}
\input{sections/Method}

\section{Experiments}
\input{sections/Experiments}

\section{Limitations and Reproducibility}
\input{sections/Limitations}

\section{Conclusion}
\input{sections/Conclusion}

\bibliographystyle{IEEEtran}
\bibliography{references}

\end{document}

%% file: sections/Abstract.tex
\begin{abstract}
A robot sent to a named gas leak must preserve gas identity, estimate the
source, and navigate to the resulting goal. We present SmellDiffusion, a
simulation pipeline that represents species-specific gas zones in an
open-vocabulary olfactory scene graph and shares the selected goal between
classical and diffusion planners. Its key components are a peak-local
geometric gate for selective source correction and diffusion-based,
gas-guided trajectory generation. Among 424 unique source--wind
configurations in solved flow, 28 have a concentration peak displaced more
than \SI{0.5}{m} from the source. A source-independent geometric gate,
calibrated only on the training split and evaluated at the observed peak,
detects 9 of 10 held-out displacements at 0.64 precision. Gating a
precomputed forward-matching correction reduces mean error on the displaced
cases from \SI{1.468}{m} to \SI{0.592}{m} (60\%), using matching for only
14/204 cases. All-case mean error falls from \SI{0.205}{m} to
\SI{0.180}{m}. All planners receive the same scene-graph source estimate as
their goal. In a controlled comparison, best-of-ten diffusion achieves mean
gas exposure comparable to gas-guided A$^\star$ (0.0476 versus 0.0455). A
single diffusion proposal takes \SI{41.7}{ms}, compared with
\SI{72.3}{ms} for gas-guided A$^\star$, although best-of-ten sequential
sampling increases total runtime. Plain A$^\star$ also reaches the same goal
and remains the fastest and shortest-path method. Six matched Gazebo runs
give mean robot-to-source errors of \SI{0.39}{m} for A$^\star$ and
\SI{0.31}{m} for diffusion.
\end{abstract}

%% file: sections/Introduction.tex
Mobile robot olfaction supports leak inspection, hazardous-material response,
and environmental monitoring~\cite{gslreview}. Unlike visual perception, olfactory sensing is intermittent: gas transport is strongly shaped by airflow and obstacles, so a measurement carries neither an object boundary nor an immediate direction to its source. A multi-gas deployment adds a semantic requirement: an operator must
be able to request a particular leak rather than follow whichever plume is
strongest.

Open-vocabulary scene graphs provide a useful abstraction for the latter
problem by associating spatial nodes with language embeddings
~\cite{conceptgraphs,hovsg}. Their usual visual features, however, do not
identify an invisible chemical species. 

We present SmellDiffusion, which instead constructs an olfactory scene graph from species-specific virtual-sensor maps: each gas zone stores a text embedding and a source estimate, and a natural-language query selects the estimate that becomes the navigation goal of a quadruped robot. The graph chooses \emph{where}; either A$^\star$ or a diffusion policy chooses \emph{how} to reach it.

\begin{figure*}[t]
\centering
\includegraphics[width=\textwidth]{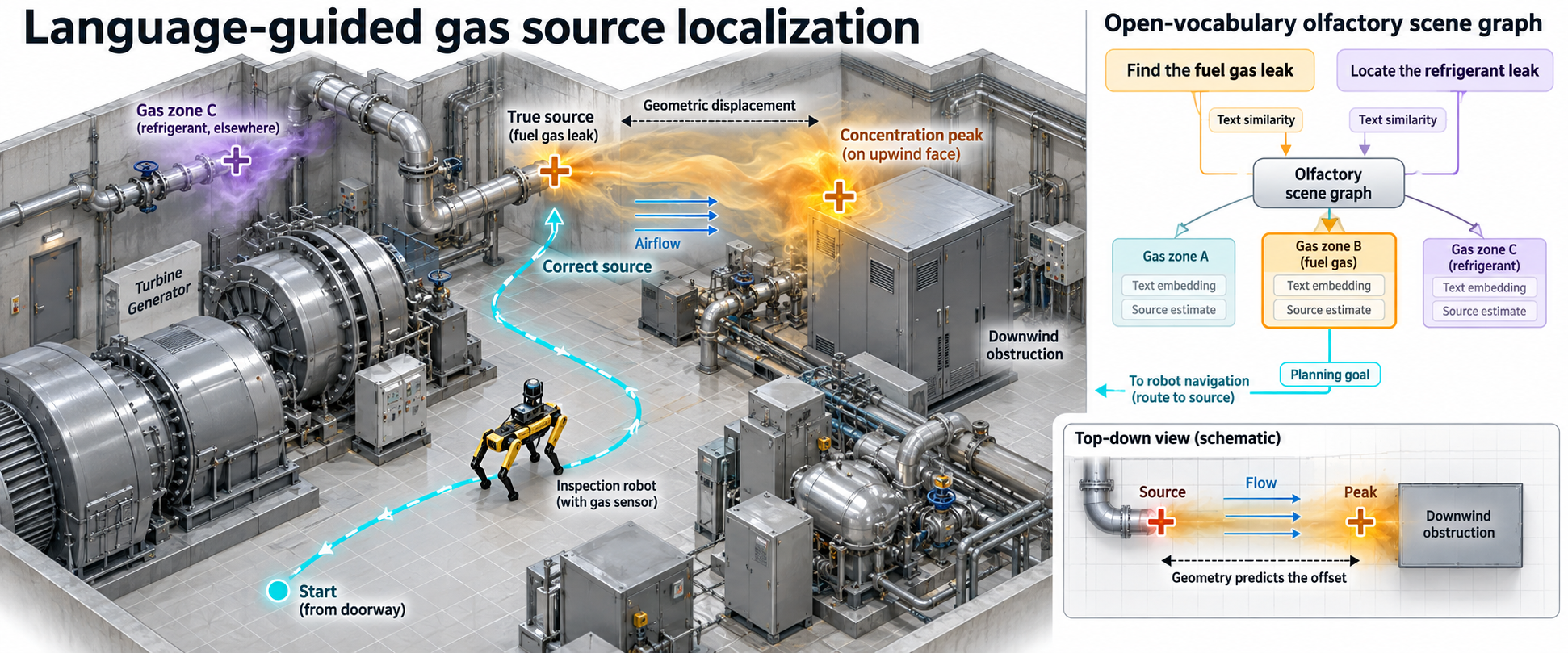}
\caption{\textbf{Conceptual overview.} A language query selects a
species-specific zone and its source estimate, which becomes the robot's
goal. The industrial environment, robot, and stylized plumes are conceptual
artwork; the right-hand graph and lower-right geometry panel are schematic.
Artwork provenance is disclosed in the acknowledgment.}
\label{fig:teaser}
\end{figure*}

Two coupled problems determine the usefulness of this representation.
Obstacle-induced flow can displace the concentration maximum from the source,
so a gas-specific goal needs a correction decision as well as an estimate.
Once the goal is selected, the route generator must use the available gas
evidence while remaining computationally practical. SmellDiffusion addresses
these problems through selective geometric correction and gas-guided
trajectory generation (Fig.~\ref{fig:teaser}). Our contributions are:

\begin{itemize}
  \item \textbf{Gas-specific language-to-goal grounding:} an olfactory scene
  graph retains species identity and links a query to a stored source
  estimate, providing a common goal interface for classical and diffusion
  planning;
  \item \textbf{Selective geometric source correction:} a training-calibrated
  gate uses flow and geometry at the observed peak, reducing held-out
  displaced-case error by \textbf{60\%} while invoking forward matching for
  only \textbf{6.9\%} of evaluation cases; and
  \item \textbf{Gas-guided diffusion planning:} with both A$^\star$
  variants and diffusion given the same source-estimate goal, best-of-ten
  diffusion achieves mean gas exposure comparable to gas-guided A$^\star$
  (0.0476 versus 0.0455), while a single diffusion proposal has lower
  measured planning time than gas-guided A$^\star$ (\SI{41.7}{ms} versus
  \SI{72.3}{ms}). Plain A$^\star$ remains fastest overall. In six matched
  Gazebo runs, mean robot-to-source error is \SI{0.39}{m} for A$^\star$
  and \SI{0.31}{m} for diffusion.
\end{itemize}

The simulation study separates representation, source estimation, and route
generation so that each contribution can be measured. Offline virtual
sensing supplies species-specific maps, and Gazebo pose supports repeatable
execution. Section~VIII details the scope and deployment considerations.

%% file: sections/Related_work.tex
\textbf{Gas source estimation and search.} Reactive plume tracing and
Infotaxis couple sensing to action~\cite{plumetrace,infotaxis}. Park
\emph{et al.} use a Gaussian mixture model to select continuous Infotaxis
actions and validate the approach in UAV flight experiments~\cite{park_gmm};
patchiness-aware multi-robot search and vision--olfaction fusion address
other real-world failure modes~\cite{sniffysquad,visolf}. We use an offline
sweep instead of active search. GADEN supplies obstacle-aware filament
dispersion through a precomputed CFD field~\cite{gaden}. Most closely related,
Ojeda \emph{et al.} compare measurements with online dispersion simulations
for candidate sources~\cite{ojeda}; our inexpensive library lookup is a
coordinate-based implementation of forward matching. Our contribution is
the peak-local decision that selectively invokes correction.

\textbf{Language-grounded spatial representations.} ConceptGraphs and
HOV-SG enrich geometric maps with open-vocabulary features for language-driven
navigation~\cite{conceptgraphs,hovsg}. We attach CLIP text embeddings to gas
zones because gasses have no visual appearance~\cite{clip}. This uses CLIP in a text-to-text setting rather than the image--text setting it was trained for, so we explicitly test held-out queries and a hubness correction ~\cite{hubs,nnn}.

\textbf{Diffusion planning.} Diffuser conditions trajectories by inpainting
known states during denoising~\cite{diffuser,ddpm}. Motion Planning Diffusion,
potential-based guidance, PRESTO, and DiPPeST demonstrate scene-conditioned or
guided diffusion for motion planning~\cite{mpd,potential,presto,dippest}. Our
policy generates 2-D waypoint sequences, inpaints the requested endpoints,
and optionally follows a clearance gradient. We retain A$^\star$ both as the
classical comparator and as a safety repair, and report when that repair is
needed. Building on these planning methods, SmellDiffusion couples
gas-specific language goals with plume-aware route selection and quantifies
how inference-time guidance reduces the subsequent repair workload.

%% file: sections/Method.tex
\begin{figure*}[t]
\centering
\includegraphics[width=0.98\textwidth]{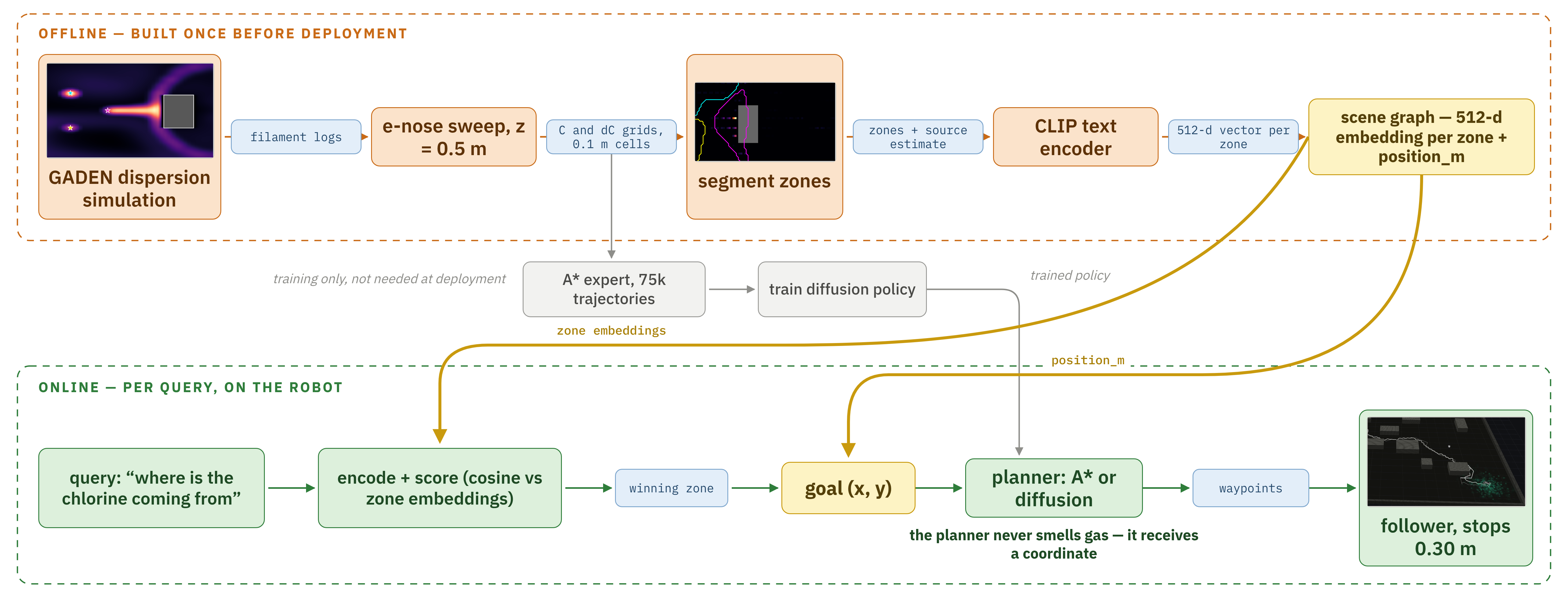}
\caption{ 
Offline virtual sensing produces per-gas grids, zones, source estimates, and text embeddings; online query matching selects a stored goal, which A$^\star$ or the diffusion policy turns into a route. Diffusion proposals are projected to free space and repaired with A$^\star$ before execution.}
\label{fig:pipeline}
\end{figure*}

\subsection{Offline graph construction}

Separate GADEN simulations provide each species' concentration field. A
virtual electronic nose follows a serpentine path at $z=\SI{0.5}{m}$ and
accumulates measurements on a \SI{0.1}{m} grid. Gaussian smoothing and adaptive thresholding
partition detections into zones; each zone stores a peak-based source
estimate and a 512-dimensional CLIP embedding of a sentence that names and
describes the gas. Environment, Zone, and Source nodes retain geometry,
species identity, and position. Because the simulator supplies a separate
channel per gas, this stage isolates representation and estimation under
known species identity; physical mixture separation is a subsequent
hardware-stage requirement.


\subsection{Online query and execution}

A query is embedded by the same CLIP text encoder and scored against zone
descriptions. The selected zone's stored position becomes a common goal for
A$^\star$ or diffusion. Diffusion proposals are projected away from occupied
cells, and every intersecting segment is replaced by an A$^\star$ subpath.
A pure-pursuit controller follows the resulting route in Gazebo using
simulator ground-truth pose. Thus the expert planner is absent from raw
diffusion sampling but remains part of deployed safety post-processing. The system generates and tracks routes in a known static map; reactive collision avoidance is outside the evaluated scope.

 Source-localization experiments use OpenFOAM-derived flow; the larger planning arenas use uniform wind, so they neither model obstacle wakes nor support displacement claims. The virtual sweep and graph are built before navigation. The online
run evaluates query-to-goal selection and route execution; active sensing
while the quadruped walks is the next integration stage.

\subsection{Trajectory representation and denoising}

Let $X_0\in\mathbb{R}^{32\times2}$ denote an expert waypoint sequence,
normalized to $[-1,1]^2$ using the room width and height. The condition
$h=(s,g,M)$ contains the normalized start, goal, and three map channels
$M=[C,A,O]$: concentration, an auxiliary gas channel, and occupancy.
Training follows the noise-prediction objective~\cite{ddpm,diffuser},
\begin{align}
 X_t &= \sqrt{\bar\alpha_t}X_0+
          \sqrt{1-\bar\alpha_t}\,\epsilon,\\
 \mathcal{L}(\theta)&=\mathbb{E}_{X_0,t,\epsilon}
   \bigl[\|\epsilon-\epsilon_\theta(X_t,t,h)\|_F^2\bigr],
\end{align}
where $\epsilon\sim\mathcal{N}(0,I)$,
$\bar\alpha_t=\prod_{j=1}^{t}(1-\beta_j)$, and 64 linearly spaced
$\beta_j$ range from $10^{-4}$ to $0.02$. At inference, Gaussian noise is
iteratively denoised; the first and last waypoints are overwritten with
$s$ and $g$ before each step and after the final step. Small endpoint error
therefore comes from endpoint conditioning. We assess segment validity and
robot execution with separate measurements.

The v1 denoiser pools its $3\times60\times100$ map into one condition
vector and processes the flattened trajectory with three MLP residual
blocks. The v3 model instead resamples the full room to
$3\times64\times64$, bilinearly reads all three channels at each noisy
waypoint, and supplies these local features to six temporal convolutional
residual blocks. The v1 planner comparison and the v3 guidance ablation therefore evaluate different saved policies and are not rows of one architecture-controlled benchmark.

Clearance guidance acts after each denoising update. It adds a scaled
finite-difference clearance gradient where obstacle distance is below
\SI{0.9}{m}, converts the displacement from world coordinates back to the
normalized frame, and then adds the scheduled diffusion noise. The gradient
is computed per grid cell in the implementation; consequently, scale 5 is
calibrated for the evaluated map resolution.
Endpoint inpainting is reapplied at the next step. Neither this local
gradient nor the learned map features certify the connecting segments.

The planner is trained separately on A$^\star$ trajectories. We use two saved policies. The original policy (v1) is an MLP denoising diffusion model with 64 denoising steps and 32 two-dimensional waypoints, trained in a 10 m$\times$6 m frame. A spatially resolved policy (v3) samples map conditioning at each waypoint; its checkpoint metadata records 75,000 procedural A$^\star$ trajectories over 250 layouts. Both clamp start and goal throughout denoising. The controlled planner comparison uses v1; the guidance ablation uses v3.

\section{Obstacle-Aware Source Estimation}
\label{sec:source}

\subsection{Peak accuracy and field displacement}

Reconstruction from simulator filament logs separates estimator error from
field displacement. In the reference ethanol case, the true concentration
peak is \SI{2.30}{m} from the source and the posterior estimate is
\SI{2.40}{m} away: the estimator tracks the peak to \SI{0.10}{m}, revealing
field displacement as the dominant source-error component. For methane and
chlorine, the field peaks near their sources while the fixed-height sweep
captures a limited portion of the rising or sinking plume. Matching the
estimator to that sparse grid is therefore more valuable
than increasing posterior complexity (Table~\ref{tab:est}). Field
displacement is reported as a reference for peak-following methods.

\begin{table}[t]
\caption{Source-estimation error from the same multi-gas sweep. Field-peak
displacement is shown as a reference, not an estimator-wide lower bound.}
\label{tab:est}
\centering
\setlength{\tabcolsep}{3.5pt}\footnotesize
\begin{tabular}{lccc}
\toprule
error (m) $\downarrow$ & methane & chlorine & CO \\
\midrule
plume posterior & 6.18 & 9.29 & 1.36 \\
zone centroid & 1.16 & 0.27 & \textbf{1.16} \\
smoothed peak & \textbf{0.10} & \textbf{0.22} & 1.22 \\
field-peak displacement & 0.07 & 0.07 & 1.25 \\
\bottomrule
\end{tabular}
\end{table}

\subsection{Deduplicated geometry study}

The saved sweep has 452 rows but 424 unique source-position/wind
configurations: 28 rows are exact duplicates. We remove identical duplicates
before analysis and preserve the position-hash split, yielding 220 training
and 204 evaluation configurations with no source position shared across
splits.  We define a configuration as displaced when the peak-to-source distance exceeds 0.5 m. Displacement occurs in 28/424 unique cases:
 0/40 at the \SI{0.1}{m/s} inlet, 0/192 at
\SI{0.5}{m/s}, and 28/192 at \SI{1.0}{m/s}; conclusions therefore cover
these three inlet settings.

Flow around a surface-mounted obstacle motivates an explanatory geometric
model~\cite{bluffbody_flow}. Let $\ell$ be the exact ray distance from the
\emph{true source} to the CAD obstacle along the local flow. We fit

 $$\widehat d = \ell-c, \qquad c=\SI{0.497}{m}$$

on the 12 displaced training configurations whose source ray intersects the obstacle. Under the same eligibility rule, the 9 held-out cases give 0.265 m mean absolute error (MAE) and $R^2=0.832$, versus 0.538 m MAE for the training-mean baseline. Seven displaced cases have
no source-ray/CAD intersection and are outside this model. Because $\ell$ and
the flow here are evaluated at the true source, this fit explains the saved
simulation but is not itself a deployable correction.

\begin{figure}[t]
\centering
\includegraphics[width=\columnwidth]{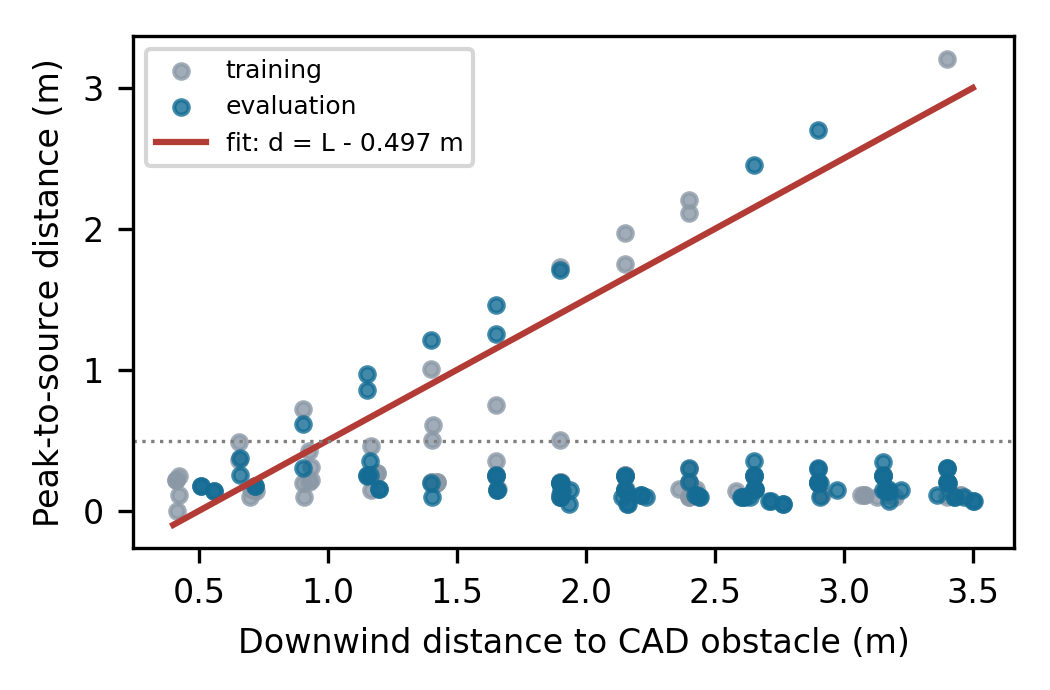}
\caption{Peak displacement versus exact source-referenced distance for the
198 unique configurations whose ray intersects the CAD obstacle. The line is
fit only to 12 displaced training cases and evaluated on nine eligible
held-out cases; non-displaced cases show the model's intended trigger domain.}
\label{fig:fit}
\end{figure}

\subsection{Peak-local gate and forward matching}
\label{sec:correction}

For a source-independent decision, we cast the ray from the \emph{observed
peak} and sample the supplied CFD flow map there. Training F1 over a fixed
grid of distance and speed thresholds selects $\ell_{\max}=\SI{1.0}{m}$ and
$u_{\min}=\SI{0.35}{m/s}$ (training F1 0.629). Distance is searched in
\SI{0.25}{m} increments and speed in \SI{0.05}{m/s} increments; flow is the
mean of CFD samples within \SI{0.4}{m} of the peak. No threshold is selected on the evaluation split.
On 204 held-out configurations the gate fires 14 times: 9 true and 5 false
positives, with one missed displacement (precision 0.64, recall 0.90).

When the gate fires, forward matching retrieves the training simulation at
the same inlet setting whose peak is nearest the observed peak and returns
its source. Forty evaluation cases have tied nearest peaks; equidistant
sources are averaged, avoiding file-order dependence.
This library lookup is a controlled proxy for simulation-per-hypothesis
matching~\cite{ojeda}. Applying it everywhere improves the ten displaced
cases while increasing errors for many already-close estimates. Gating
concentrates the benefit into 14 calls (Table~\ref{tab:correction}).

For completeness, write $p$ for the measured peak and $v$ for the inlet
setting. Let $\ell_p$ be the first CAD intersection along the peak-local
mean flow $\bar u_p$, with $\ell_p=\infty$ for a ray miss. The deployed-input
decision and the retrieval set are
\begin{align}
 G(p)&=\mathbf{1}\{\ell_p\leq\ell_{\max},\;
                         \|\bar u_p\|\geq u_{\min}\},\\
 J(p,v)&=\underset{j\in\mathcal{D}_{\rm train}:v_j=v}
                 {\operatorname{arg\,min}}\;\|p_j-p\|_2,\\
 \hat s(p,v)&=\begin{cases}
 |J|^{-1}\sum_{j\in J}s_j,&G(p)=1,\\
 p,&G(p)=0.
 \end{cases}
\end{align}
Here $s_j$ and $p_j$ are the known source and resulting peak of a training
simulation. Source coordinates are thus labels in the offline library;
the unknown evaluation source is never a query input. Numerically, peak
distances within $10^{-12}$\,m of the minimum share the retrieval set.
Exact ray intersection uses the original rectangular CAD obstacle and an \SI{8}{m} search limit.

\begin{figure*}[t]
\centering
\includegraphics[width=0.97\textwidth]{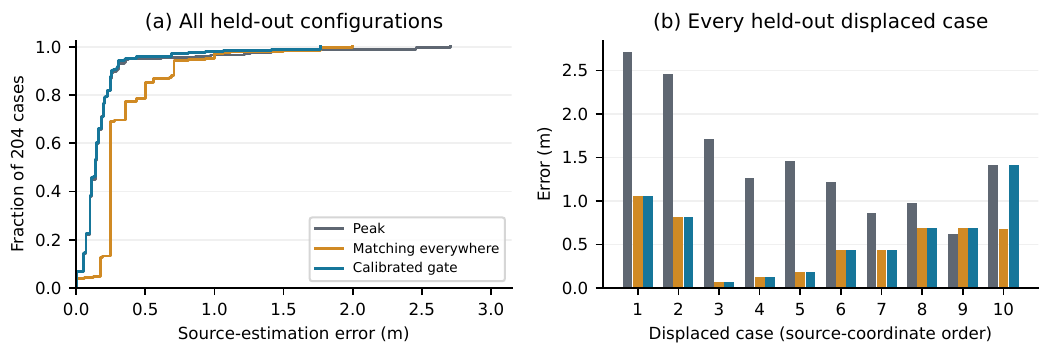}
\caption{Held-out correction outcomes from saved per-case records.
(a) Empirical cumulative source-error distributions for all 204 cases;
matching every peak improves the displaced subset while shifting the overall
distribution. (b) All ten displaced cases, ordered by source coordinates,
show the heterogeneous effect of calibrated gating. The uncorrected case
is the gate's one missed displacement. All displaced cases are shown.}
\label{fig:correction}
\end{figure*}

\subsection{Correction Influence}

Figure~\ref{fig:correction} exposes the cost of retrieval ambiguity. Using
a \SI{0.05}{m} change threshold, ungated matching improves 20 evaluation cases
and degrades 144; calibrated gating improves nine and degrades five. Among the 194
non-displaced cases, mean error is \SI{0.140}{m} for the peak,
\SI{0.348}{m} for ungated matching, and \SI{0.159}{m} for calibrated
gating. The all-case gain therefore combines a substantial improvement in
rare displaced cases with a smaller mean change in the majority group.
The gate's 189 true negatives retain the original peak. Its five false
positives are not free, and high displacement recall alone would conceal
their cost.

Nearest-peak matching also compresses an entire field into one coordinate.
Different sources can produce equal or nearby peaks, especially at an
obstacle face. Averaging tied source labels makes the estimator deterministic.
A denser library or matching full concentration and flow fields offers a
direct path to resolving the remaining inverse ambiguity.

\begin{table}[t]
\caption{Held-out source error after peak retrieval. The calibrated gate is
fit on training data; ported applies the earlier source-local thresholds
to peak-local variables without refitting.}
\label{tab:correction}
\centering
\setlength{\tabcolsep}{2.8pt}\footnotesize
\begin{tabular}{lcccc}
\toprule
method & all mean & P95 & displaced & calls \\
& \multicolumn{3}{c}{error (m) $\downarrow$} & /204 \\
\midrule
peak alone & 0.205 & 0.377 & 1.468 & 0 \\
matching everywhere & 0.356 & 0.919 & \textbf{0.517} & 204 \\
ported gate & 0.222 & 0.826 & 1.468 & 11 \\
calibrated gate & \textbf{0.180} & \textbf{0.353} & 0.592 & 14 \\
\bottomrule
\end{tabular}
\end{table}

The gate needs a mapped or estimated flow vector at the observed peak; our
experiment uses CFD rather than a real anemometer. In three additional solved
rooms, only 5/620 configurations are displaced, compared with 5/80 under
uniform sampling in the reference room (one-sided Fisher $p=0.0027$). This
localizes the effect to confined flow impinging on geometry and motivates a
broader multiroom displacement benchmark.

\section{Language Grounding}
\label{sec:lang}

The integrated pipeline's recorded query test scores 7/9. Separately, a
frozen enriched-description probe scores 8/9 on the development queries used
to select those descriptions, but only 3/10 on ten subsequently written
queries; bare gas names also score 3/10. A fixed description can win even for
gas-free reference text, consistent with embedding hubness
~\cite{hubs}. Nearest Neighbor Normalization using all 12 reference probes
($k=12$, $\alpha=1$) raises the held-out result to 5/10 without retraining
~\cite{nnn}. This ten-query set provides an initial evaluation of
training-free bias correction. Absent-gas queries use a separate symbolic
vocabulary check.

The pipeline's air-contrastive score subtracts $\cos(q,\text{air})$ from
every candidate for a fixed query. This changes an acceptance threshold but
cannot change candidate ranking. Candidate-specific normalization, as in the
NNN result, is the relevant correction.

More explicitly, for unit-normalized text embeddings the corrected score is
\begin{equation}
 S(q,z)=q^\top z-\frac{1}{12}\sum_{r\in\mathcal{R}}r^\top z,
\end{equation}
where $\mathcal{R}$ is the fixed set of gas-free reference probes. The
subtracted term varies with candidate $z$, so it can change the winner.
For example, the held-out swimming-pool query changes from air to chlorine,
and the balloon-filling query changes from gasoline vapor to helium.
The kitchen-stove and closed-garage exhaust queries still select gasoline
vapor incorrectly. These recorded successes and failures distinguish a
useful bias correction from reliable semantic source selection.

\begin{table}[t]
\caption{Language evaluation sets remain separate. The description probe
uses eight candidates; development queries informed description selection.}
\label{tab:language}
\centering\footnotesize
\begin{tabular}{llc}
\toprule
evaluation & representation & correct / total\\
\midrule
integrated query test & pipeline descriptions & 7/9\\
probe development & enriched descriptions & 8/9\\
probe held-out & bare names & 3/10\\
probe held-out & enriched descriptions & 3/10\\
probe held-out & enriched + NNN & 5/10\\
\bottomrule
\end{tabular}
\end{table}

Species-aware maps do make the selected coordinate query-dependent. With
separate maps, smoothed-peak errors for methane, chlorine, and carbon monoxide
are 0.10, 0.22, and \SI{1.22}{m}; summing the fields gives one coordinate for
all queries and errors of 2.34, 2.51, and \SI{1.22}{m}. The mean rises from
0.51 to \SI{2.02}{m}. This ablation establishes the value of retaining gas
identity, not the necessity of a graph: a flat species-to-position dictionary
was not tested.

%% file: sections/Experiments.tex
\label{sec:planner}

\subsection{Protocol}

We sample 12 free start--goal pairs at least \SI{3}{m} apart with NumPy seed
0 in the existing \SI{10}{m}$\times$\SI{6}{m} map. Plain A$^\star$,
gas-guided A$^\star$, one v1 diffusion proposal, and best-of-ten diffusion
use a common concentration-field source and occupancy map. Diffusion receives
a zero auxiliary channel. Its input adapter applies $\log(1+C)$ and
renormalizes, while gas-guided A$^\star$ uses normalized $C$ directly; the
information source is shared, but the tensors are not identical. The neural
occupancy channel is uninflated; all methods use the same inflated free
space for path validity. The gas-guided A$^\star$ cost uses fixed weight 2. The
diffusion arms use the deployed clearance-guidance scale 5. The best-of-ten
arm selects maximum mean exposure among segment-valid raw proposals;
the one-sample arm is the first of those same ten draws.

All paths are sampled every \SI{0.05}{m} for exposure and clearance. Validity
checks continuous line segments against inflated obstacles rather than only
waypoints. Success requires a valid path to the planning goal. Runtime uses
one CPU Torch thread after an untimed warm-up; fixed seeds control diffusion
sampling. Means and sample standard deviations cover successful paths, while
success counts include every request.

\subsection{Costs, metrics, and validity}

Both A$^\star$ variants use eight-connected search without corner cutting.
Their edge cost is grid-step length times
$1+0.5\max(0,1-D/0.60)+w_g(1-C)$, where $D$ is clearance in meters and
$w_g$ is 0 or 2. Greedy line-of-sight shortcutting follows search; a shortcut
rejected by continuous validation falls back to the unshortened grid path.
Occupancy inflation rounds the \SI{0.25}{m} robot radius up to three cells
on the \SI{0.1}{m} map. Reported clearance is distance to the original
occupied cells, rather than residual distance beyond the inflated boundary.

For a path $P=(x_1,\ldots,x_K)$, length is
$L(P)=\sum_{i=1}^{K-1}\|x_{i+1}-x_i\|_2$. Let $\tilde x_j$ be the
common arc-length samples, including the last endpoint. We report
\begin{equation}
 E(P)=\frac{1}{N}\sum_{j=1}^N C(\tilde x_j),\qquad
 D_{\min}(P)=\min_j D(\tilde x_j).
\end{equation}
Exposure is a dimensionless mean normalized concentration, not inhaled
dose or accumulated physical gas. Maximizing it tests plume-seeking route
selection; a gas-avoidance mission would require a different objective.
Collision validity is evaluated independently of these samples by traversing
every grid interval crossed by each segment, checking both sides at cell
corners. Thus fine waypoint spacing cannot hide a collision between points.

Runtime covers A$^\star$ search and shortcutting or diffusion denoising,
including guidance. Map preparation, common collision/quality evaluation,
fallback after a rejected A$^\star$ shortcut, and robot execution are outside
the timer. These are planner-computation measurements on the recorded CPU
run, not end-to-end latency. Every planner request is retained in the result
CSV, including its success flag. The ten-proposal selector filters invalid
candidates, but the artifact retains the first and selected proposals rather
than all ten intermediate draws; it therefore supports per-request failure
accounting, not a retrospective raw-sample failure rate.

\begin{table*}[t]
\caption{Controlled planner comparison (mean $\pm$ s.d., 12/12 successes for
every row). Best-of-ten time includes ten sequential samples.}
\label{tab:planner}
\centering
\setlength{\tabcolsep}{8pt}\small
\begin{tabular}{lcccc}
\toprule
planner & exposure $\uparrow$ & clearance (m) $\uparrow$ & length (m) $\downarrow$ & runtime (ms) $\downarrow$ \\
\midrule
A$^\star$ & 0.0302$\pm$0.0431 & 0.442$\pm$0.097 & \textbf{5.76}$\pm$1.62 & \textbf{15.2}$\pm$8.9 \\
gas A$^\star$ & 0.0455$\pm$0.0688 & 0.446$\pm$0.088 & 5.78$\pm$1.61 & 72.3$\pm$32.6 \\
guided diff., 1 & 0.0313$\pm$0.0405 & \textbf{0.589}$\pm$0.150 & 8.00$\pm$2.14 & 41.7$\pm$1.8 \\
guided diff., best 10 & \textbf{0.0476}$\pm$0.0427 & 0.547$\pm$0.114 & 9.71$\pm$3.46 & 407.7$\pm$9.9 \\
\bottomrule
\end{tabular}
\end{table*}

\begin{figure*}[t]
\centering
\includegraphics[width=0.97\textwidth]{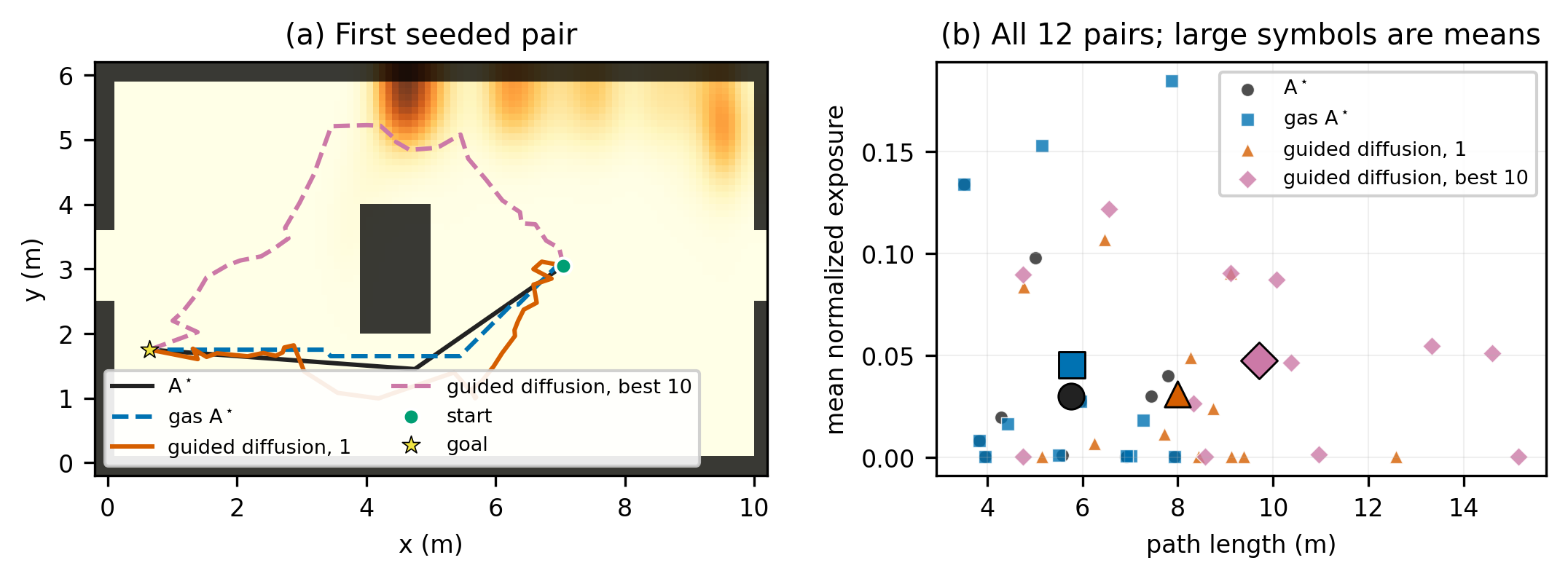}
\caption{Corrected common-input planner evaluation. (a) The first pair in
the seeded generation order, shown without outcome-based selection; the
background is normalized concentration and dark cells are inflated
obstacles. Curves are segment-valid raw paths before deployed projection or
repair. (b) Length--exposure outcomes for all 12 pairs; enlarged symbols are
method means. Best-of-ten selects exposure and consequently need not select
the shortest or clearest diffusion sample.}
\label{fig:planner}
\end{figure*}

\subsection{Results and clearance guidance}

Every method succeeds on all 12 requests (Table~\ref{tab:planner} and
Fig.~\ref{fig:planner}). A$^\star$ produces the shortest paths and has the
lowest mean planning time: $15.2\pm8.9$~ms, compared with
$41.7\pm1.8$~ms for one guided diffusion sample and
$407.7\pm9.9$~ms for ten sequential samples. A single diffusion proposal
has higher minimum clearance but is 39\% longer; selecting ten proposals
for gas exposure raises exposure to 0.0476, comparable to gas-guided
A$^\star$ at 0.0455, while increasing length and latency.
\textbf{Diffusion's advantage is controllable route diversity:} one
sample increases mean minimum clearance over A$^\star$ by 33\%, while
selecting among ten samples raises exposure by 52\% over a single draw.
This lets one learned sampler express clearance- and plume-oriented routes
without changing the goal. A$^\star$ retains the length and runtime
advantage. Independent diffusion samples can also be batched.

We separately evaluate inference-time clearance guidance with the same v3
checkpoint, guidance scales 0 and 5, and ten paired sampling seeds.
\textbf{In the cluttered scene, guidance cuts segments requiring A$^\star$
repair from $7.7\pm1.95$ to $1.8\pm1.23$ of 31 (24.8\% to 5.8\%), a 77\%
relative reduction, and cuts projected waypoints from $12.9\pm2.02$ to
$1.9\pm1.10$ of 32 (85\%).} In the scattered scene, repaired segments fall
from $2.2\pm0.79$ to $1.3\pm0.67$ (7.1\% to 4.2\%), while projected waypoints
fall from $7.3\pm0.48$ to $2.9\pm1.97$ (60\%). The two scenes were used in
earlier guidance tuning, so this experiment measures the mechanism's paired
effect in the targeted operating conditions. Projection and A$^\star$ repair
provide the final execution safeguard.

\input{figures/guidance_table.tex}

Table~\ref{tab:guidance} includes both checkpoints and both scenes. Each
entry pairs the same ten seeds across guidance scales. Repair percentages use
the original 31 segments after waypoint projection, before adding any
A$^\star$ subpaths. This ablation uses the deployed grid line-of-sight repair
test, sampled at approximately half-cell spacing; the controlled planner
comparison supplies the complementary continuous segment check. Together
they measure deployed repair workload and raw-path validity.

\section{Integrated Gazebo Demonstrations}

The assembled pipeline is run with A$^\star$ and diffusion for each gas in an
authored \SI{24}{m}$\times$\SI{16}{m} Gazebo arena. Each run starts from a
fresh spawn at the same commanded pose and evaluates one requested species;
the settled initial poses differ by at most \SI{0.13}{m} within a gas pair.
The straight path to each requested goal is obstructed. Table~\ref{tab:runs}
reports source-estimation, arrival, and route quantities separately.

\begin{table}[t]
\caption{Matched Gazebo execution results. Arrival error is measured from the
final simulator pose to the ground-truth source; one run is recorded per gas
and planner.}
\label{tab:runs}
\centering
\setlength{\tabcolsep}{3.5pt}\footnotesize
\begin{tabular}{lccc}
\toprule
& chlorine & ethanol & methane \\
\midrule
source-estimate error (m) & 0.16 & 0.07 & 0.25 \\
A$^\star$ arrival error (m) & 0.41 & 0.28 & 0.49 \\
diffusion arrival error (m) & \textbf{0.24} & \textbf{0.25} & \textbf{0.44} \\
\bottomrule
\end{tabular}
\end{table}

The follower uses a \SI{0.30}{m} stopping tolerance around the planned goal.
A$^\star$ records robot-to-source errors of 0.41, 0.28, and \SI{0.49}{m};
diffusion records 0.24, 0.25, and \SI{0.44}{m}. The corresponding descriptive
means are $0.39\pm0.11$ and $0.31\pm0.11$~m (sample s.d.). Diffusion is lower
for all three gases, but one execution per condition does not support an
inferential performance claim. Simulator pose isolates upstream errors but
does not predict how SLAM error would combine with them.

\subsection{Separating estimation, goal projection, and arrival}

Let $s^\star$ be the physical source location, $\hat s$ the stored source
estimate, $g$ the collision-free planning goal after any projection, and
$x_f$ the final robot position. The triangle inequality gives
\begin{equation}
 \|x_f-s^\star\|\leq
 \underbrace{\|x_f-g\|}_{\text{following}}+
 \underbrace{\|g-\hat s\|}_{\text{goal projection}}+
 \underbrace{\|\hat s-s^\star\|}_{\text{estimation}}.
\end{equation}
The follower tolerance controls only the first term. Projection can be
necessary when a concentration peak lies beside an obstacle, even if the
estimate itself is accurate. These terms may partly cancel geometrically,
so their sum is a bound rather than the measured arrival error reported in
Table~\ref{tab:runs}. A physical deployment would additionally need to account
for pose-estimation error and the chosen robot reference point.

The demonstrations test query-to-motion execution in a uniform-wind arena;
the geometry benchmark tests source correction in solved obstacle flow.
Together they evaluate complementary links of the complete pipeline.

%% file: figures/guidance_table.tex
\begin{table}[t]
\caption{Guidance ablation: original segments requiring repair (\%, mean $\pm$ s.d., ten seeds). Scales 0 and 5 use the same checkpoint and paired random seeds.}
\label{tab:guidance}
\centering\footnotesize
\begin{tabular}{llcc}
\toprule
scene & model & scale 0 & scale 5 \\
\midrule
scattered & v1 & 8.39$\pm$2.26 & 3.23$\pm$2.63 \\
scattered & v3 & 7.10$\pm$2.54 & 4.19$\pm$2.18 \\
cluttered & v1 & 27.74$\pm$8.36 & 7.42$\pm$3.74 \\
cluttered & v3 & \textbf{24.84$\pm$6.28} & \textbf{5.81$\pm$3.97} \\
\bottomrule
\end{tabular}
\end{table}

%% file: sections/Limitations.tex
\begin{table*}[t]
\caption{Scope of the available evidence. Counts refer to the experimental
unit in each row and must not be pooled into a single success rate.}
\label{tab:evidence}
\centering\footnotesize
\setlength{\tabcolsep}{5pt}
\begin{tabular}{p{0.19\textwidth}p{0.22\textwidth}p{0.49\textwidth}}
\toprule
study & experimental units & supported interpretation\\
\midrule
reference-room geometry & 424 unique configurations & 220 training / 204 evaluation; 28 displacements, including 10 in evaluation.\\
source-referenced fit & 12 training / 9 evaluation & Eligible displaced ray hits; explanatory fit uses true-source information.\\
peak-local correction & 204 evaluation cases & Training-calibrated gate invokes retrieval in 14 cases; same-inlet library.\\
additional solved rooms & 620 configurations & Only five displacements; insufficient events for a transferable fit.\\
language probe & 9 development / 10 held-out & Description selection and subsequent query testing are separate.\\
planner comparison & 12 pairs / four methods & Planning time and static-map route metrics under fixed inputs and seeds.\\
guidance ablation & 2 scenes / 2 models / 10 seeds & Paired scales 0 and 5 on reused scenes; deployed repair workload.\\
Gazebo comparison & 3 gases / 2 planners & Same nominal spawn and controller; one execution per condition.\\
\bottomrule
\end{tabular}
\end{table*}

All evidence is simulation-based. The fixed \SI{0.5}{m} sensing plane is
poorly matched to buoyant or dense gases; separate simulator channels bypass
real mixture identification; and the flow needed by the geometric gate comes
from CFD. The position-hash split is retained from prior work and the
evaluation set was already examined before this revision. Although no revised
threshold was chosen on its outcomes, it is not a new prospective test. The
ten-query language set and two guidance scenes are exploratory. The diffusion
pipeline still depends on projection and A$^\star$ repair, and no physical
quadruped experiment is reported.

Future work will replace the fixed sensing plane and species-separated
simulator channels with onboard mixture-aware sensing, estimate flow online,
and evaluate the complete system on a physical quadruped under localization
uncertainty and changing plumes. It will also test the correction gate and
diffusion planner prospectively across unseen geometries, start--goal pairs,
and repeated executions before drawing general performance conclusions.

The revision artifact fixes all random seeds and contains the deduplication,
exact ray--box intersection, peak-local flow lookup, per-case matching output,
continuous segment checker, common \SI{0.05}{m} path sampler, and input
SHA-256 hashes. It reports exclusions and request success explicitly. The
procedural multi-layout training set can be regenerated from code, but the
large array itself is not included; the checkpoint metadata records its layout, trajectory, step, and architecture counts. As clarified in Table~\ref{tab:evidence}, planner standard deviations cover 12 requests and guidance standard deviations cover ten draws at fixed endpoints; runtime remains hardware dependent.

%% file: sections/Conclusion.tex
SmellDiffusion connects natural-language gas requests to quadruped routes
through gas-specific scene-graph goals, selective geometric source correction,
and guided diffusion planning. \textbf{A training-calibrated peak-local gate
detects 9/10 held-out displacements, reduces their mean error from
\SI{1.468}{m} to \SI{0.592}{m}, and invokes forward matching for only 14/204
cases.} Diffusion provides a complementary planning advantage: one guided
sample increases mean minimum clearance over A$^\star$ by \textbf{33\%},
best-of-ten selection increases exposure over one sample by \textbf{52\%},
and guidance reduces cluttered-scene repair demand by \textbf{77\%}.
A$^\star$ remains the faster, shorter-path option and supplies the execution
safeguard. Together, these results establish a complete, reproducible
language-to-motion system and clear gains from selective correction and
guided trajectory generation.

%% file: references.bib
@article{gslreview,
  title   = {Gas source localization and mapping with mobile robots: A review},
  author  = {Francis, Adam and Li, Shuai and Griffiths, Christian and Sienz, Johann},
  journal = {Journal of Field Robotics},
  volume  = {39},
  number  = {8},
  pages   = {1341--1373},
  year    = {2022},
  publisher = {Wiley Online Library},
  doi     = {10.1002/rob.22109}
}

@article{ojeda,
  author  = {Ojeda, Pepe and Monroy, Javier and Gonzalez-Jimenez, Javier},
  title   = {Robotic Gas Source Localization With Probabilistic Mapping and
             Online Dispersion Simulation},
  journal = {IEEE Transactions on Robotics},
  volume  = {40},
  pages   = {3551--3564},
  year    = {2024},
  doi     = {10.1109/TRO.2024.3426368}
}

@inproceedings{hovsg,
  title     = {Hierarchical Open-Vocabulary 3D Scene Graphs for Language-Grounded
               Robot Navigation},
  author    = {Werby, Abdelrhman and Huang, Chenguang and B{\"u}chner, Martin and
               Valada, Abhinav and Burgard, Wolfram},
  booktitle = {Proceedings of Robotics: Science and Systems},
  address   = {Delft, Netherlands},
  year      = {2024},
  doi       = {10.15607/RSS.2024.XX.077}
}

@inproceedings{dippest,
  author    = {Stamatopoulou, Maria and Liu, Jianwei and Kanoulas, Dimitrios},
  booktitle = {Proc. IEEE/RSJ Int. Conf. on Intelligent Robots and Systems (IROS)},
  title     = {{DiPPeST}: Diffusion-based Path Planner for Synthesizing
               Trajectories Applied on Quadruped Robots},
  year      = {2024},
  pages     = {7787--7793},
  doi       = {10.1109/IROS58592.2024.10802677}
}

@article{gaden,
  author  = {Monroy, Javier and Hernandez-Bennetts, Victor and Fan, Han and
             Lilienthal, Achim and Gonzalez-Jimenez, Javier},
  title   = {{GADEN}: A 3D Gas Dispersion Simulator for Mobile Robot Olfaction
             in Realistic Environments},
  journal = {Sensors},
  volume  = {17},
  number  = {7},
  pages   = {1479},
  year    = {2017},
  doi     = {10.3390/s17071479}
}

@article{infotaxis,
  author  = {Vergassola, Massimo and Villermaux, Emmanuel and Shraiman, Boris I.},
  title   = {`Infotaxis' as a strategy for searching without gradients},
  journal = {Nature},
  volume  = {445},
  number  = {7126},
  pages   = {406--409},
  year    = {2007},
  doi     = {10.1038/nature05464}
}

@article{park_gmm,
  author  = {Park, Minkyu and An, Seulbi and Seo, Jaemin and Oh, Hyondong},
  title   = {Autonomous Source Search for {UAV}s Using Gaussian Mixture
             Model-Based Infotaxis: Algorithm and Flight Experiments},
  journal = {IEEE Transactions on Aerospace and Electronic Systems},
  volume  = {57},
  number  = {6},
  pages   = {4238--4254},
  year    = {2021},
  doi     = {10.1109/TAES.2021.3098132},
  note    = {IEEE document 9492004}
}

@article{bluffbody_flow,
  author  = {Martinuzzi, Robert and Tropea, Cameron},
  title   = {The Flow Around Surface-Mounted, Prismatic Obstacles Placed in a
             Fully Developed Channel Flow (Data Bank Contribution)},
  journal = {Journal of Fluids Engineering},
  volume  = {115},
  number  = {1},
  pages   = {85--92},
  year    = {1993},
  doi     = {10.1115/1.2910118}
}

@inproceedings{clip,
  author    = {Radford, Alec and Kim, Jong Wook and Hallacy, Chris and
               Ramesh, Aditya and Goh, Gabriel and Agarwal, Sandhini and
               Sastry, Girish and Askell, Amanda and Mishkin, Pamela and
               Clark, Jack and Krueger, Gretchen and Sutskever, Ilya},
  title     = {Learning Transferable Visual Models From Natural Language
               Supervision},
  booktitle = {Proc. Int. Conf. Machine Learning (ICML)},
  series    = {PMLR},
  volume    = {139},
  pages     = {8748--8763},
  year      = {2021}
}

@article{hubs,
  author  = {Radovanovi\'{c}, Milo{\v{s}} and Nanopoulos, Alexandros and
             Ivanovi\'{c}, Mirjana},
  title   = {Hubs in Space: Popular Nearest Neighbors in High-Dimensional Data},
  journal = {Journal of Machine Learning Research},
  volume  = {11},
  pages   = {2487--2531},
  year    = {2010}
}

@inproceedings{nnn,
  author    = {Chowdhury, Neil and Wang, Franklin and Shenoy, Sumedh and
               Kiela, Douwe and Schwettmann, Sarah and Thrush, Tristan},
  title     = {Nearest Neighbor Normalization Improves Multimodal Retrieval},
  booktitle = {Proc. Conf. Empirical Methods in Natural Language Processing
               (EMNLP)},
  pages     = {22571--22582},
  year      = {2024},
  doi       = {10.18653/v1/2024.emnlp-main.1257}
}

@inproceedings{mpd,
  author    = {Carvalho, Joao and Le, An T. and Baierl, Mark and
               Koert, Dorothea and Peters, Jan},
  title     = {Motion Planning Diffusion: Learning and Planning of Robot
               Motions with Diffusion Models},
  booktitle = {Proc. IEEE/RSJ Int. Conf. on Intelligent Robots and Systems
               (IROS)},
  pages     = {1916--1923},
  year      = {2023},
  doi       = {10.1109/IROS55552.2023.10342382}
}

@inproceedings{potential,
  author    = {Luo, Yunhao and Sun, Chen and Tenenbaum, Joshua B. and Du, Yilun},
  title     = {Potential Based Diffusion Motion Planning},
  booktitle = {Proc. Int. Conf. Machine Learning (ICML)},
  series    = {PMLR},
  volume    = {235},
  pages     = {33486--33510},
  year      = {2024}
}

@inproceedings{presto,
  author    = {Seo, Mingyo and Cho, Yoonyoung and Sung, Yoonchang and
               Stone, Peter and Zhu, Yuke and Kim, Beomjoon},
  title     = {{PRESTO}: Fast Motion Planning Using Diffusion Models Based on
               Key-Configuration Environment Representation},
  booktitle = {Proc. IEEE Int. Conf. Robotics and Automation (ICRA)},
  pages     = {10861--10867},
  year      = {2025},
  doi       = {10.1109/ICRA55743.2025.11128590}
}

@inproceedings{ddpm,
  author    = {Ho, Jonathan and Jain, Ajay and Abbeel, Pieter},
  title     = {Denoising Diffusion Probabilistic Models},
  booktitle = {Advances in Neural Information Processing Systems (NeurIPS)},
  volume    = {33},
  pages     = {6840--6851},
  year      = {2020}
}

@inproceedings{diffuser,
  author    = {Janner, Michael and Du, Yilun and Tenenbaum, Joshua B. and
               Levine, Sergey},
  title     = {Planning with Diffusion for Flexible Behavior Synthesis},
  booktitle = {Proc. Int. Conf. Machine Learning (ICML)},
  series    = {PMLR},
  volume    = {162},
  pages     = {9902--9915},
  year      = {2022}
}

@inproceedings{conceptgraphs,
  author    = {Gu, Qiao and Kuwajerwala, Alihusein and Morin, Sacha and others},
  title     = {{ConceptGraphs}: Open-Vocabulary 3D Scene Graphs for Perception
               and Planning},
  booktitle = {Proc. IEEE Int. Conf. Robotics and Automation (ICRA)},
  pages     = {5021--5028},
  year      = {2024},
  doi       = {10.1109/ICRA57147.2024.10610243}
}

@article{plumetrace,
  author  = {Li, Wei and Farrell, Jay A. and Card{\'e}, Ring T.},
  title   = {Tracking of Fluid-Advected Odor Plumes: Strategies Inspired by
             Insect Orientation to Pheromone},
  journal = {Adaptive Behavior},
  volume  = {9},
  number  = {3--4},
  pages   = {143--170},
  year    = {2001},
  doi     = {10.1177/10597123010093003}
}

@article{visolf,
  author  = {Hassan, Sunzid and Wang, Lingxiao and Mahmud, Khan Raqib},
  title   = {Robotic Odor Source Localization via Vision and Olfaction Fusion
             Navigation Algorithm},
  journal = {Sensors},
  volume  = {24},
  number  = {7},
  pages   = {2309},
  year    = {2024},
  doi     = {10.3390/s24072309}
}

@article{sniffysquad,
  author  = {Cheng, Yuhan and Chen, Xuecheng and Yang, Yixuan and Wang,
             Haoyang and Xu, Jingao and Hong, Chaopeng and Zhang, Xiao-Ping
             and Liu, Yunhao and Chen, Xinlei},
  title   = {{SniffySquad}: Patchiness-Aware Gas Source Localization with
             Multi-Robot Collaboration},
  journal = {arXiv preprint arXiv:2411.06121},
  year    = {2024},
  url     = {https://arxiv.org/abs/2411.06121}
}
